\documentclass[11pt]{article}

\usepackage[utf8]{inputenc}
\usepackage[T1]{fontenc}
\usepackage{lmodern}
\usepackage{microtype}

\usepackage[margin=1in]{geometry}

\usepackage{times}
\usepackage{helvet}
\usepackage{courier}

\usepackage[hyphens]{url}
\usepackage{graphicx}
\usepackage{xcolor}
\usepackage{multirow}
\usepackage[numbers]{natbib}
\usepackage{caption}
\usepackage{csquotes}
\usepackage{booktabs}
\usepackage{amsmath,amssymb}
\usepackage{authblk}
\usepackage{tcolorbox}
\usepackage{hyperref}

\title{Semantic Uncertainty-Guided Orchestration in Hierarchical Multi-Agent Systems}

\author[1]{John Knowlton}
\author[2]{Aritra Guha}
\author[1,3]{Risto Miikkulainen}

\affil[1]{The University of Texas at Austin, USA}
\affil[2]{AT\&T Chief Data Office, USA}
\affil[3]{Cognizant AI Lab, USA}

\date{}

\begin{document}

\maketitle

\begin{abstract}
As large language model (LLM)-based multi-agent systems become increasingly capable, coordinating agents under uncertainty becomes a fundamental challenge. Existing orchestration strategies typically rely on fixed interaction patterns and often lack mechanisms for assessing the reliability of intermediate reasoning steps, allowing errors and hallucinations to propagate through the system. This paper introduces a semantic-uncertainty-guided orchestration approach, HASSUM as a general framework for uncertainty-aware coordination in multi-agent systems. The method estimates uncertainty using semantic entropy and semantic density, which measure trust at the level of answer semantics rather than output probabilities. These signals enable adaptive orchestration decisions, including output verification, selective reprompting, additional deliberation, and confidence-aware response selection. Because the approach operates independently of any particular agent architecture, it can be integrated into a broad range of hierarchical and collaborative multi-agent systems. The evaluations demonstrate an implementation within a hierarchical agent framework and evaluate it on StrategyQA, JailbreakBench, and TruthfulQA benchmarks. Across tasks that require complex reasoning and are prone to ambiguity or hallucinations, uncertainty-guided orchestration yields more reliable outcomes than uncertainty-unaware coordination. Semantic entropy and semantic density in tandem outperformed either metric alone. Ablations testing different thresholds and model sizes demonstrated that both influence the effectiveness of semantic metrics. The results suggest that semantic uncertainty is a practical and general-purpose signal for improving robustness and trustworthiness in agentic AI systems.
\end{abstract}

\section{Introduction}

Multi-Agent Systems (MASs) enable structured decomposition and coordination that single Large Language Models (LLMs) often struggle to provide, and have shown promise for long-horizon reasoning, tool use, and autonomous task execution~\citep{li2023camelcommunicativeagentsmind, park2023generativeagentsinteractivesimulacra}. However, this added complexity introduces new failure modes: agents may hallucinate, reason inconsistently, or propagate errors across intermediate steps~\citep{gu2025agentgroupchatv2divideandconquerllmbasedmultiagent}. Existing mitigations such as self-consistency~\citep{wang2023selfconsistencyimproveschainthought} improve output quality through heuristic sampling, but treat semantically equivalent responses as distinct whenever their surface form differs, leaving a gap between lexical variation and genuine disagreement in meaning.

Semantic uncertainty methods address this gap by quantifying uncertainty over meaning rather than tokens. Semantic entropy~\citep{kuhn2023semanticuncertaintylinguisticinvariances} measures disagreement across discrete clusters of meaning, while semantic density~\citep{qiu2024semanticdensityuncertaintyquantification} measures the continuous concentration of responses in embedding space. The two capture different failure signatures --- entropy is sensitive to contradiction, density to imprecision --- making their combination complementary rather than redundant. Both metrics were originally developed and validated as standalone measures of single-model output reliability; whether either can function as a \emph{decision-making} signal inside a coordinating multi-agent system, determining not just how uncertain a response is but what the system should do about it, has not been studied.

Recent work has demonstrated that uncertainty estimates can influence agent behavior, including inference-time filtering, diagnostic handoff, ensemble reweighting, and adaptive collaboration. However, existing approaches typically rely on a single uncertainty signal and rarely examine how uncertainty can guide orchestration decisions within hierarchical multi-agent systems. More importantly, little is known about when uncertainty-driven intervention actually improves performance and when it instead introduces unnecessary computation or degrades reasoning quality. This leaves open the broader question of whether semantic uncertainty can serve not merely as an evaluation metric, but as an actionable control signal for agent orchestration.

This gap motivates two questions. First, can multiple semantic uncertainty metrics function as a live control signal inside a hierarchical multi-agent system rather than a post-hoc score --- that is, can a central orchestrator use them to decide when to reprompt, reassign, or delegate, rather than simply to flag or filter completed outputs? Second, when does this kind of uncertainty-driven intervention actually help, and when does it not? We answer both by embedding semantic entropy and density into the orchestration loop of HASHIRU~\citep{pai2025hashiruhierarchicalagenthybrid}, a hierarchical MAS in which a CEO agent allocates subtasks to specialized worker agents and integrates their outputs. The resulting system, Hierarchical Agent System with Semantic Uncertainty Metrics (HASSUM), uses the combined signal to actively branch execution --- triggering reprompting, worker reassignment, or additional delegation before the CEO commits to a final answer.

Our contributions are threefold. First, we introduce HASSUM, a semantic-uncertainty-guided orchestration framework that uses semantic entropy and semantic density as live control signals for hierarchical multi-agent decision making. Second, we demonstrate that semantic entropy and semantic density capture complementary failure modes, and that their combination outperforms either metric alone when used for orchestration. Third, through evaluation across nine benchmarks, we characterize a boundary condition for uncertainty-guided orchestration: semantic uncertainty is most beneficial when failures arise from ambiguity or semantic instability, but provides limited benefit when failures stem from factual gaps, evidence retrieval failures, or logical errors.

We evaluate HASSUM on StrategyQA~\citep{geva2021strategyqa}, JailbreakBench~\citep{chao2024jailbreakbenchopenrobustnessbenchmark}, and TruthfulQA~\citep{lin2022truthfulqameasuringmodelsmimic} --- benchmarks selected for their susceptibility to ambiguity and hallucination --- and six additional benchmarks outside the primary ambiguity domain. Semantic uncertainty improves performance on ambiguity- and hallucination-driven tasks but provides limited or negative benefit when failures arise from retrieval errors, factual gaps, or logical mistakes. These results suggest that the value of uncertainty-guided orchestration depends strongly on the dominant failure mode.

\section{Related Work}
\label{sec:related}

A growing body of work has explored uncertainty as a mechanism for shaping agent behavior rather than merely evaluating outputs after execution. Existing approaches fall into three broad categories: (i) uncertainty-guided control and orchestration, where uncertainty determines which action an agent should take next; (ii) adaptive collaboration methods, which modify communication or interaction patterns among agents; and (iii) uncertainty estimation methods that quantify confidence but do not alter execution. HASSUM sits at the intersection of these lines of research by using semantic uncertainty to drive orchestration decisions within a hierarchical multi-agent system. \citet{stoisser2025uncertainty} guide inference-time filtering and abstention in a single structured-reasoning agent using retrieval and summary uncertainty; \citet{srivastava2026agentic} gate diagnostic handoff in a multi-agent healthcare system using semantic entropy alone; \citet{sun2026coe} decompose multi-LLM uncertainty into intra- and inter-model components to reweight ensemble outputs. \citet{wang2026guided} similarly use entropy-based signals to dynamically guide agent behavior in heterogeneous multi-agent systems, but target a flat strong-weak collaboration setting rather than a hierarchical controller. Each relies on a single uncertainty signal within a flat architecture --- a single agent, a gate on one agent's output, or an ensemble reweighting scheme --- rather than a hierarchical controller that branches execution across discrete actions, and none characterize when such intervention helps versus fails.

Adaptive orchestration work asks a related but distinct question: when is additional agent interaction worthwhile, independent of semantic uncertainty. DOWN~\citep{eo2025debatenecessaryadaptivemultiagent} gates structured debate using token-level or verbalized confidence while introducing a new collaboration protocol; MAD~\citep{liang2024encouragingdivergentthinkinglarge} redesigns inter-agent communication itself through peer debate judged by a third agent; \citet{zhao2026doesmultiagentcollaborationhelp} take a purely diagnostic stance, characterizing when collaboration succeeds or fails without proposing a control mechanism. \citet{li2026rethinking} use entropy reduction as a supervisory signal to decide when tool calls are worth issuing, addressing a related but narrower question --- tool-call quality rather than worker-response reliability --- without a centralized reprompt/reassignment mechanism. Redesigning peer interaction in this way is often costly --- requiring new communication schemes, judge or debate mechanisms, and re-validation across tasks --- and such redesigns typically do not transfer cleanly to existing deployed architectures.

Uncertainty estimation has also been extended to agentic reasoning without altering execution. UProp~\citep{duan2025upropinvestigatinguncertaintypropagation} decomposes trajectory-level uncertainty into intrinsic and propagated components via pointwise mutual information, and SAUP~\citep{zhao2024saupsituationawarenessuncertainty} propagates uncertainty across ReAct-style reasoning steps with situational awareness; both improve estimation quality for sequential agents but stop at measurement, never using the resulting estimate to alter execution. \citet{jiang2026discouq} similarly stop at estimation, learning calibrated confidence from the structure of inter-agent disagreement in an ensemble rather than using the resulting estimate to alter which agent acts next.

Inference-time refinement offers a further route to reliability without a central decision point. Self-consistency~\citep{wang2023selfconsistencyimproveschainthought} samples multiple reasoning paths and selects the most consistent answer, but applies the same fixed sampling budget regardless of whether an output is actually uncertain. Archon~\citep{saadfalcon2025archonarchitecturesearchframework} searches over refinement pipelines combining sampling, ensembling, ranking, critique, and verification, but the pipeline is fixed at design time rather than adapted per query. DebUnc~\citep{yoffe2025debuncimprovinglargelanguage} incorporates uncertainty into peer-to-peer debate, but decision-making is distributed across peers rather than resolved by a single point of control. \citet{wang2026t2po} instead uses uncertainty to control exploration during RL training itself, resampling low-information turns rather than intervening at inference time.

Unlike prior work, our goal is not to introduce a new uncertainty estimator. Instead, we investigate uncertainty as a control signal for hierarchical orchestration. HASSUM uses semantic entropy and semantic density to decide when additional reasoning effort should be allocated through reprompting, worker reassignment, or delegation, and studies the conditions under which such interventions succeed or fail. This emphasis on recoverability and intervention distinguishes our work from prior uncertainty-estimation and collaboration-focused approaches.

\paragraph{Semantic Uncertainty Metrics}
\label{sec:metrics}

We use two uncertainty signals in this work, both computed from a set of stochastic completions sampled from a worker agent for the same prompt.

Semantic entropy~\citep{kuhn2023semanticuncertaintylinguisticinvariances} quantifies uncertainty over discrete clusters of meaning. Sampled responses are grouped by pairwise entailment into semantic-equivalence classes, and each cluster is assigned a probability equal to its share of the samples. Entropy is then the Shannon entropy over this cluster distribution,
\begin{equation}
    H = - \sum_{i=1}^{k} p(C_i) \log p(C_i),
\end{equation}
where $C_i$ is the $i^{th}$ semantic cluster; low entropy indicates that sampled responses converge on a single meaning.

Semantic density~\citep{qiu2024semanticdensityuncertaintyquantification} instead quantifies uncertainty over a continuous embedding space, without explicit clustering. Each response is embedded using all-MiniLM-L6-v2~\citep{wang2021minilmv2multiheadselfattentionrelation}, and density is estimated via a kernel-based formulation,
\begin{equation}
    D = \frac{1}{N^2} \sum_{i=1}^{N} \sum_{j=1}^{N} K\big(\phi(y_i), \phi(y_j)\big),
\end{equation}
using a dimensionally-invariant Epanechnikov kernel $K$; tightly clustered embeddings yield high density.

Although both metrics are derived from the same sampled responses, they are sensitive to different failure signatures. Density operates on continuous embedding geometry and is sensitive to vagueness and imprecision, flagging responses that agree in direction but lack semantic precision. Entropy operates on discrete entailment logic and is sensitive to contradiction, flagging responses that disagree in meaning even when they are topically close with highly similar embeddings. This structural difference is what makes their combination principled rather than merely additive: each metric detects a class of uncertainty the other is blind to by construction. Both are recomputed per worker response and passed to the CEO as the primary orchestration signal.

\section{Orchestration Architecture}
\label{sec:HASSUM}

HASSUM builds on HASHIRU~\citep{pai2025hashiruhierarchicalagenthybrid}, a hierarchical multi-agent system in which a CEO agent orchestrates specialized worker agents through decomposition, specialization, and coordinated reasoning. We selected HASHIRU because its CEO already performs the orchestration decisions semantic uncertainty is designed to inform --- worker selection, reprompting, and delegation --- providing a natural substrate for testing whether semantic signals can improve decisions the architecture already supports.

\begin{figure*}[t!]
\centering
\includegraphics[width=1.017\textwidth,trim={8.2ex 0ex 0ex 0ex},clip]{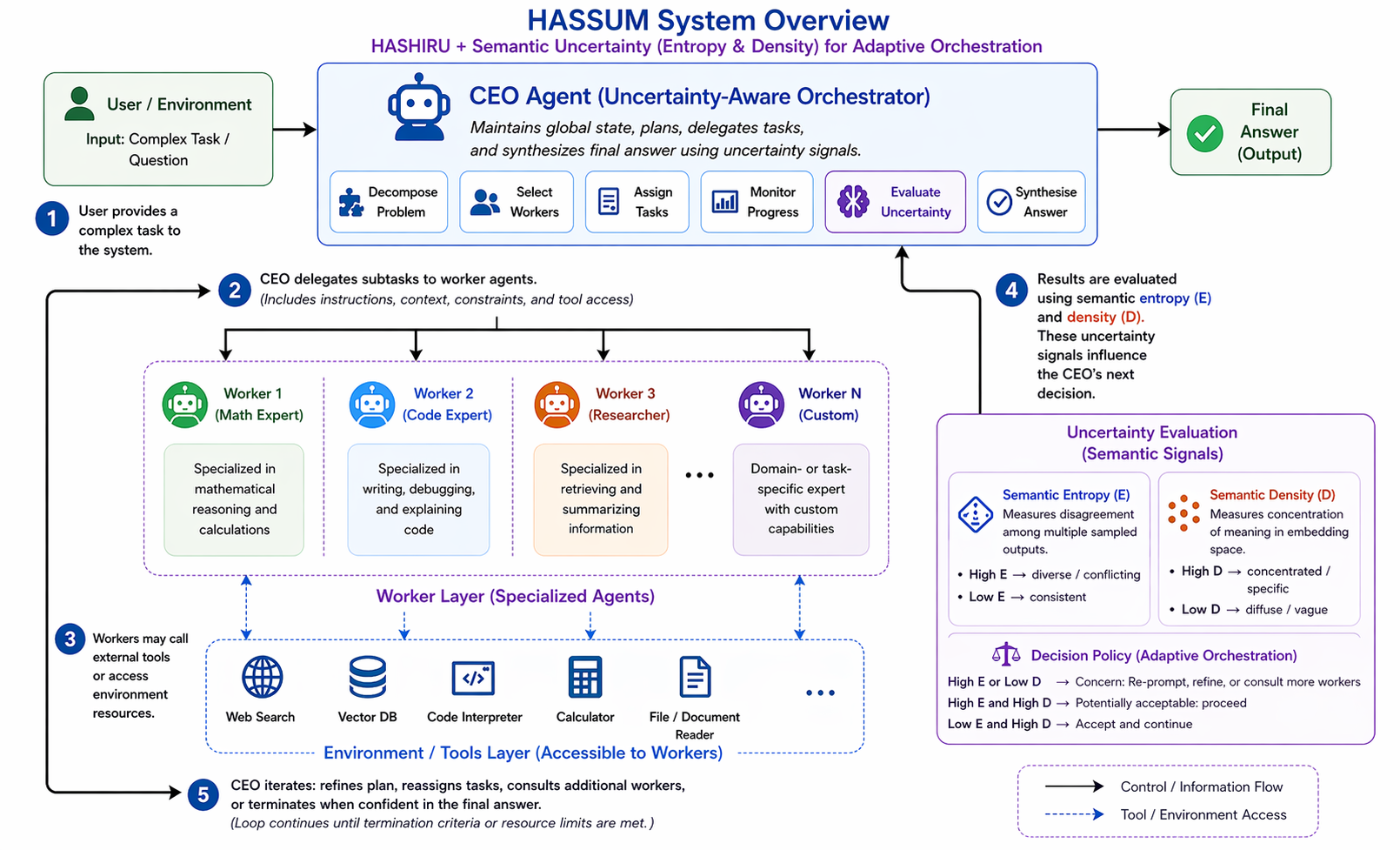}
\caption{High-level flowchart of HASSUM. The CEO agent receives the user query, selects an appropriate specialized worker agent, and evaluates the returned response using semantic density and semantic entropy. If the response exceeds the uncertainty thresholds, the CEO accepts the answer and synthesizes the final response. If semantic concern is triggered, the CEO may reprompt the same worker, select a different worker with a more suitable expertise, or invoke multiple workers in parallel before reevaluating the outputs. This process continues until the CEO reaches a sufficiently confident answer or the maximum number of reasoning rounds is reached. Thus, HASSUM transforms uncertainty estimation from a passive evaluation metric into an active control mechanism that directly guides agent selection, refinement, and final decision making.}
\label{fg:HASSUM_workflow}
\end{figure*}

The CEO interprets queries, decomposes them into subtasks, and assigns each to a specialized worker agent or external tool (e.g., web search, a calculator, or a code interpreter), ``firing'' underperforming workers and ``hiring'' new ones as needed. HASSUM's workflow is an iterative, potentially multi-turn loop, shown in Figure~\ref{fg:HASSUM_workflow}: (i) the CEO determines whether a query needs decomposition, (ii) allocates workers and tools to subtasks, (iii) workers generate candidate responses, (iv) the CEO evaluates these using semantic density and entropy and decides whether refinement is needed, and (v) once satisfied, synthesizes a final response.

At step (iv), the CEO computes density and entropy over additional stochastic completions of the worker's response and compares each scalar against a pre-set acceptance threshold (e.g., density $\geq 0.8$, entropy $\leq 1.1$ in our experiments), rather than acting on the raw value directly. Density at or above threshold together with entropy at or below threshold is accepted outright; any other combination --- either scalar falling outside its threshold --- warrants closer inspection and typically triggers refinement: a revised prompt, delegation to a different worker, or parallel consultation of multiple workers, repeating across turns until the CEO is confident or a reasoning-round cap is reached. This augments the existing harness with an external evaluation layer, without modifying the underlying language models; threshold selection and its effect on this tradeoff are examined in Section~\ref{sec:ablation}.

Since HASHIRU's CEO sometimes answers directly with minimal delegation, we added an explicit multi-agent call to elicit more hierarchical behavior; for ablations, entropy and density can each be disabled independently, and delegation can be forced. We also added JSONL trace logging and a reprompt cap to prevent unbounded refinement loops.

We additionally de-emphasize HASHIRU's resource-budget system, which trades off local compute against API calls: in preliminary experiments, active budget constraints suppressed the multi-turn, multi-agent behavior we aim to study, limiting worker counts and biasing the CEO toward tool use over delegation. Leaving this unconstrained lets us more directly observe how semantic uncertainty shapes the CEO's decisions; we leave the interaction between resource budgets and uncertainty-guided orchestration to future work.

\section{Experiments}
\label{sec:Benchmarks}

We evaluate HASSUM on three primary benchmarks selected for their susceptibility to ambiguity and hallucination, and on six additional benchmarks spanning task types that are less characteristically driven by semantic ambiguity, used to test generalization. For each benchmark, we evaluate a consecutive subset of problems starting from the first item.

\paragraph{Primary benchmarks.} StrategyQA~\citep{geva2021strategyqa} requires multi-step, implicit reasoning to answer binary yes/no questions, combining multiple pieces of background knowledge through intermediate reasoning steps not explicitly stated in the prompt --- a task well suited to MASs, since different agents can contribute partial reasoning steps or intermediate facts. We use 500 of the total 2780 questions. An example question is:
\begin{displayquote}
    \emph{Did Jon Brower Minnoch suffer from anorexia nervosa?}
\end{displayquote}
This question appears without context, but to illustrate what it tests: John Brower Minnoch is the heaviest recorded human in history, and anorexia nervosa is an eating disorder characterized by an intense fear of gaining weight.

JailbreakBench~\citep{chao2024jailbreakbenchopenrobustnessbenchmark} evaluates robustness to adversarial prompts that rephrase or disguise harmful intent to bypass safety constraints, scored by refusal behavior as a binary safe/unsafe signal. We use all 200 questions.

TruthfulQA~\citep{lin2022truthfulqameasuringmodelsmimic} evaluates a model's tendency to produce truthful versus misleading answers, using questions that induce hallucinations or reflect common misconceptions where incorrect answers are often plausible or widely believed. Responses are scored against reference answers as truthful, partially truthful, or false. We use 500 of the total 817 questions.

\paragraph{Scope benchmarks.} To test whether the benefits observed on these three benchmarks generalize, we additionally evaluate on MuSiQue and HotpotQA (multi-hop question answering), GSM8K (mathematical reasoning), MMLU-Pro Law (expert multiple-choice reasoning), MuSR (long-form deductive reasoning), and IFBench (instruction following). Results on these benchmarks are discussed in Section~\ref{sec:scope} alongside the primary results.

\subsection{Experimental Setup}

Each worker response is supplemented with four additional stochastic samples at temperature 0.85 to compute semantic density and semantic entropy. Workers were typically DeepSeek-R1 (8.2B parameters) or Llama-3.2 (3.2B parameters), chosen as small models capable of running locally; HASHIRU supports substantially larger models as well. The CEO was typically Gemini 2.0 Flash, matching HASHIRU's original experimental setup; this model was retired around June 2026, so the final experiments were run with Gemini 2.5 Flash instead.

Since our focus is the effect of semantic density and semantic entropy rather than model choice, results should not depend heavily on which models are used. To verify this, one ablation instead used GPT-5.4, a substantially larger model, as both worker and CEO (Section~\ref{sec:ablation}).

HASSUM was run using a combination of local computation and API calls: smaller models such as Llama-3.2 ran locally via Ollama (v0.18.3), while larger models such as Gemini 2.0 Flash were queried via API. Experiments were conducted on a personal Windows machine using WSL2 Ubuntu 22.04, with 16GB of RAM, an Intel(R) Core(TM) i5-10400F CPU, and an NVIDIA GeForce RTX 3060.

\section{Results}
\label{sec:results}

HASSUM was evaluated across two benchmark groups: three primary benchmarks with high ambiguity, and six additional benchmarks to evaluate the scope of the method. Ablation studies focused on semantic metrics, thresholds, and worker model capability.

\subsection{Main Findings}

\begin{table}[t!]
\centering
\caption{Comparison of HASSUM and HASHIRU, i.e.\ multi-agent orchestration with and without semantic uncertainty metrics. Accuracy is percentage correct with bootstrapped confidence intervals; the numbers in parentheses indicate data subset size.}
\label{tab:results}
\begin{tabular}{lcc}
\toprule
\textbf{Benchmark} & \textbf{HASSUM} & \textbf{HASHIRU} \\
 & \textbf{(Metrics On)} & \textbf{(Metrics Off)} \\
\midrule
\multicolumn{3}{c}{\textit{Benchmarks with high ambiguity}} \\
\midrule
StrategyQA (500) & $78\% \pm 2.7\%$ & $70\% \pm 4.0\%$ \\
TruthfulQA (500) & $41\% \pm 3.9\%$ & $19\% \pm 4.8\%$ \\
JailbreakBench & $61\% \pm 6.3\%$ & $25\% \pm 7.6\%$ \\
\midrule
\multicolumn{3}{c}{\textit{Benchmarks outside the scope}} \\
\midrule
MuSR (500) & $74\% \pm 5.3\%$ & $73\% \pm 4.6\%$\\
MuSiQue (500) & $45\% \pm 4.8\%$ & $55\% \pm 4.9\%$\\
HotpotQA (500) & $54\% \pm 3.5\%$ & $68\% \pm 3.2\%$\\
IFBench (100) & $55\% \pm 6.7\%$ & $47\% \pm 5.4\%$ \\
GSM8K (100) & $93\% \pm 5.7\%$ & $96\% \pm 5.4\%$ \\
MMLU-Pro Law (100) & $55\% \pm 7.2\%$ & $56\% \pm 7.1\%$ \\
\bottomrule
\end{tabular}
\end{table}

Across the three primary benchmarks, semantic uncertainty consistently improved the CEO's ability to recognize unreliable intermediate reasoning and adapt its orchestration strategy, rather than simply increasing total reasoning effort (Table~\ref{tab:results}). The gains scale inversely with baseline reliability: JailbreakBench and TruthfulQA, where HASHIRU's unaided accuracy was lowest (25\% and 19\% respectively), saw the largest proportional improvements --- accuracy more than doubled on both, with non-overlapping confidence intervals in each case. StrategyQA,  the baseline was already comparatively strong (70\%) and the gain was smaller, but still eight percentage points. This pattern suggests that semantic uncertainty is most valuable precisely where the baseline system is least reliable, rather than providing a uniform boost regardless of starting point. As will be discussed in Section~\ref{sec:discussion}, the uncertainty metrics catch a specific class of recoverable failure, which is most prevalent when the baseline accuracy is low. This mechanistic link is what the two metrics are designed to detect, and with the distinct roles they played throughout the results.

JailbreakBench and TruthfulQA baselines are lower in large part because unaided workers frequently produce vague, hedged, or partially-committed responses rather than outright contradictions, and semantic density --- sensitive to this kind of imprecision even when responses agree in direction --- is well matched to catching it. Where sampled responses instead reached genuinely competing conclusions, semantic entropy was the metric that flagged concern, since such disagreement can occur even when individual responses each appear locally confident and dense. StrategyQA's smaller relative gain illustrates the boundary of both metrics rather than a role for either alone: its baseline failures more often stem from a single incorrect factual premise stated confidently and consistently, a pattern that is neither divergent enough to raise entropy nor imprecise enough to lower density, since the resulting samples are uniform in both position and phrasing.

\subsection{Ablation Studies}
\label{sec:ablation}

Three ablations were tested, each holding the benchmark subset and all other settings fixed while varying one factor.

\begin{figure}[t]
\centering
\includegraphics[width=0.7\linewidth]{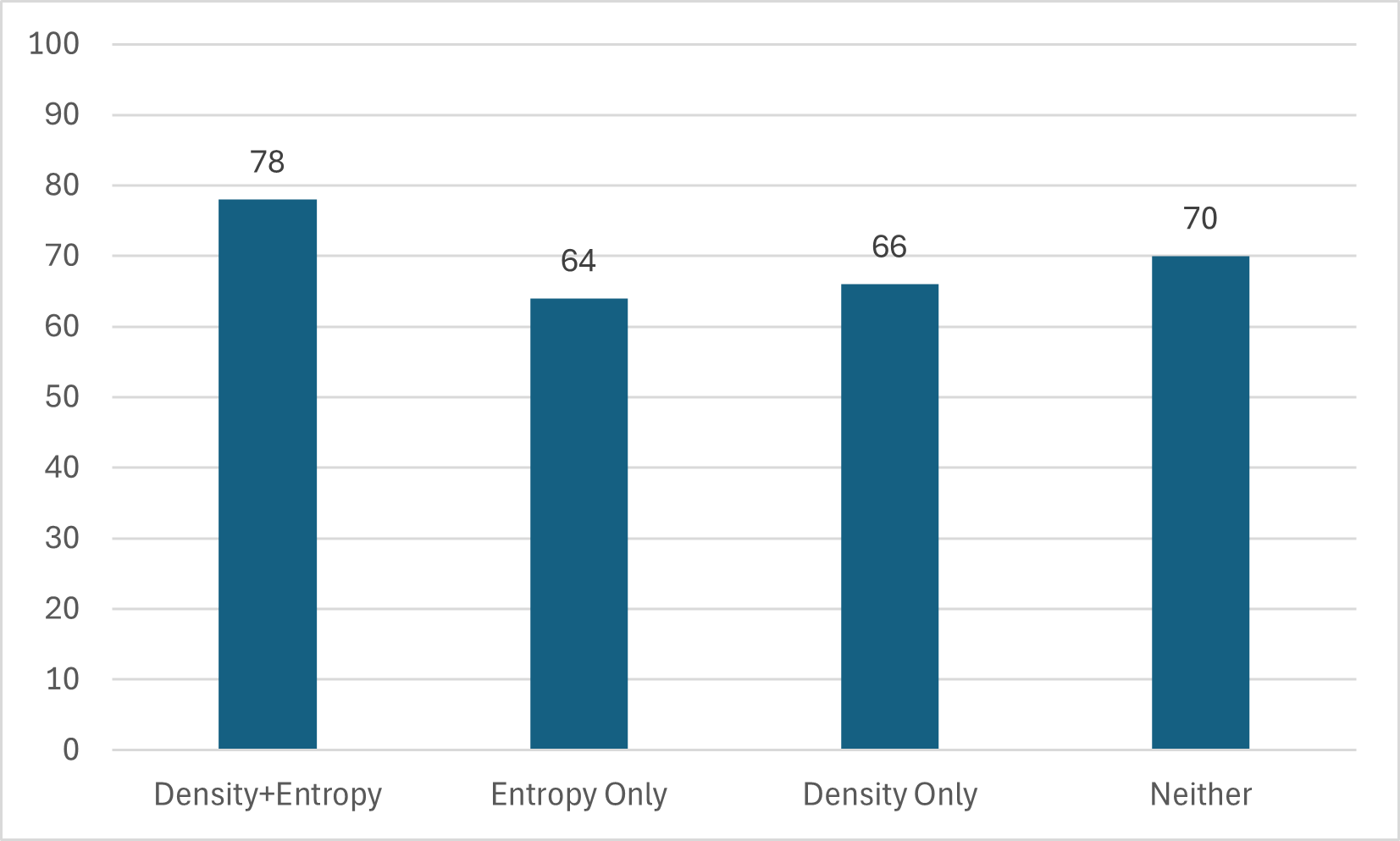}
\caption{Accuracy across ablations of semantic density and entropy on the StrategyQA subset. The combination performed best; either metric alone underperformed even the no-metric baseline.}
\label{fg:ablations_SD_SE}
\end{figure}

\textbf{Metric isolation.} HASSUM was run on the same StrategyQA subset under four conditions: density and entropy enabled together, each enabled alone, and both disabled, to test whether the metrics' contributions are additive or complementary. Figure~\ref{fg:ablations_SD_SE} shows that the combined signal outperformed every other condition, and that disabling both outperformed using either alone. This is because each metric alone can mislead the CEO in a different direction: density alone is more permissive, since it does not require entailment agreement and can pass responses that agree topically but disagree in conclusion. For instance, the density-only ablation in Appendix~\ref{appendix:A} accepted a factually wrong biographical premise once density crossed threshold, even though the underlying reasoning was never corrected. On the other hand, entropy alone is more conservative but blind to fluent, internally consistent errors. Responses that agree cluster into one low-entropy group regardless of whether that response is correct. The combination works well because each metric's failure mode is exactly the case the other is designed to catch.

\textbf{Threshold sensitivity.} Holding one metric's threshold fixed at its default (density 0.8, entropy 1.1), the other metric's acceptance threshold was swept across a range of values to test how strictness trades off against cost. Raising the density threshold and lowering the entropy threshold both make the CEO stricter, since higher density and lower entropy both indicate higher confidence; both directions improved accuracy up to a point but increased reprompting frequency and computational cost (full curves are in Appendix~\ref{appendix:B}).

\textbf{Worker capability.} Holding the benchmark subset and thresholds fixed, the local workers (Llama-3.2, DeepSeek-R1) were replaced with GPT-5.4 to test whether semantic uncertainty's value depends on worker capability rather than the metrics themselves. Table~\ref{tab:small_vs_SOTA_confidence} shows that GPT-5.4 produced much higher semantic density (0.957 vs.\ 0.887) and lower semantic entropy (0.058 vs.\ 0.949) despite resulting in the same HASSUM accuracy (78\%). Stronger workers are often consistent on the first attempt, triggering fewer interventions; with weaker models, HASSUM uses such interventions to bring the accuracy of the system to the same level as with stronger models (traces are in Appendix~\ref{appendix:C}).

\begin{figure*}[h]
\centering
\includegraphics[width=1.012\textwidth,trim={2.7ex 0ex 0ex 0ex},clip]{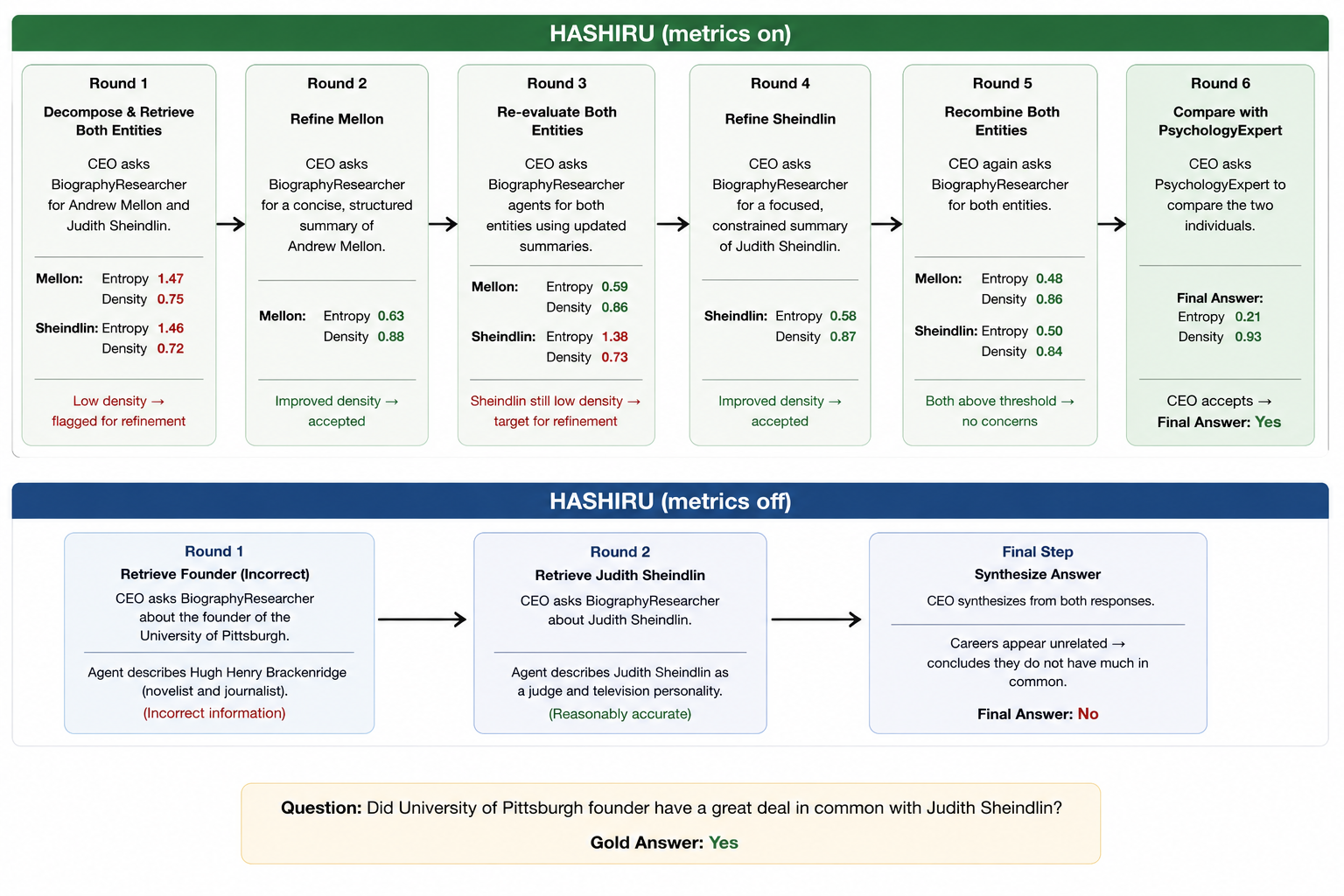}
\caption{Example execution traces on a StrategyQA question. With metrics enabled, low semantic density triggers refinement before synthesis; without metrics, the CEO accepts incorrect intermediate reasoning.}
\label{fg:Q3_Run}
\end{figure*}

\begin{table}[t]
\centering
\caption{Semantic uncertainty for local workers vs.\ GPT-5.4 on StrategyQA. Even though GPT-5.4 workers are much stronger, HASSUM can use interventions to compensate, resulting in system accuracy of 75\% in both cases.}
\setlength{\tabcolsep}{5pt}
\begin{tabular}{llcccc}
\hline
\textbf{Metric} & \textbf{Run} & \textbf{Mean} & \textbf{Median} & \textbf{Q75} \\
\hline
\multirow{2}{*}{Semantic Density}
 & Local & 0.887 & 0.883  & 0.921 \\
 & GPT-5.4 & 0.957 & 0.957  & 1.000 \\

\hline
\multirow{2}{*}{Semantic Entropy}
 & Local & 0.949 & 1.055  & 1.332 \\
 & GPT-5.4 & 0.058 & 1.7e{-16}  & 0.013 \\
\hline
\end{tabular}
\label{tab:small_vs_SOTA_confidence}
\end{table}

\subsection{Trace-Level Reprompt Analysis}

Across all three primary benchmarks, successful recoveries followed one pattern: vague or unstable worker outputs were flagged by density or entropy, prompting a reprompt, reassignment, or collaborative reasoning that resolved the ambiguity. Figure~\ref{fg:Q3_Run} illustrates this effect on a StrategyQA question asking whether the founder of the University of Pittsburgh had much in common with Judith Sheindlin. The CEO's first attempt misidentifies the founder, yielding density 0.75 and 0.72 for the two sub-answers, both below the 0.8 threshold and flagged for refinement. After several rounds of requerying and consulting a specialist agent, density rises above threshold and the CEO synthesizes the correct answer. Without the semantic metrics, the same initial misidentification is instead accepted outright, propagating the error to an incorrect final answer.

Crossing the threshold is not itself a guarantee of correction, however --- in the density-only ablation on the same question (Appendix~\ref{appendix:A}), a reprompt raised density just above threshold while leaving the underlying premise wrong. On TruthfulQA, reprompting was likewise less consistently corrective: several traces reproduced the same misconception across rounds, suggesting the CEO's default recovery (i.e.\ re-querying the same worker with modest phrasing changes) is poorly matched to persistent factual errors as opposed to incomplete or ambiguous reasoning. Thus, even though semantic concern reliably identified unstable responses on both benchmarks, it only translated into corrected answers when the recovery action matched the actual source of error.

Density was the more consistent trigger overall, catching incomplete or imprecise reasoning even when responses appeared coherent on the surface; entropy contributed specifically when sampled generations reflected genuinely competing interpretations. The two metrics thus play distinct, complementary roles rather than serving as interchangeable confidence scores.

\subsection{Evaluation of Scope}
\label{sec:scope}

To evaluate the scope of using semantic uncertainty to guide multi-agent orchestration, HASSUM was evaluated on six benchmarks outside the primary ambiguity domain: multi-hop QA (MuSiQue, HotpotQA), math (GSM8K), expert reasoning (MMLU-Pro Law), long-form deduction (MuSR), and instruction following (IFBench). As seen in Table~\ref{tab:results}, none showed consistent improvement from semantic metrics. The direction of the gap also varies in magnitude and sign across scope benchmarks in ways that track task type: the multi-hop benchmarks show the largest and most consistent losses (MuSiQue: $-10$pp, HotpotQA: $-14$pp), while IFBench shows a positive gain ($+8$pp) despite falling outside the primary ambiguity domain, and GSM8K and MMLU-Pro Law are within a few points of parity in either direction. These results suggest that the relevant distinction is not simply ``in scope'' versus ``out of scope,'' but whether a benchmark's failures are dominated by evidence-selection or reasoning-chain errors (where refinement actively hurts, as on multi-hop QA) versus constraint-satisfaction errors (where refinement can still help, as on IFBench), even when the underlying failure is not semantic ambiguity per se.

The largest gaps were on multi-hop QA, where HASHIRU outperformed HASSUM by 10 to 14 percentage points: semantic concern still triggered reprompting, but recovered few errors while flipping many correct answers, since additional reasoning mostly revisited the same incorrect trajectory. MuSR showed roughly equal positive and negative flips, netting little change despite heavy orchestration activity. On GSM8K, MMLU-Pro Law, and IFBench, failures reflected computational mistakes, knowledge gaps, or constraint violations rather than semantic instability, so incorrect responses often appeared fluent and consistent --- exactly the profile semantic uncertainty is not designed to catch.

\section{Discussion and Future Work}
\label{sec:discussion}

Semantic uncertainty is most valuable when instability is itself the source of error: recoverable failures from vague or ambiguous reasoning benefited consistently from CEO supervision, while failures rooted in factual gaps or logical error were not reliably improved regardless of how often semantic concern fired. Uncertainty is thus necessary but not sufficient for recoverability --- the central boundary condition established by our results.

This complementarity also carries an asymmetry in failure modes. A density false negative --- topically similar but contradictory responses embedding closely and passing without concern --- is the more dangerous failure, since the CEO accepts an answer masking genuine disagreement. An entropy false negative --- vague responses clustering loosely into one group --- is less harmful, since the model still converges on one direction. This suggests entropy's contribution exceeds its lower trigger frequency: it fires less often because it is a more specific detector, and catches the more consequential failure when it does. Conversely, false positives carry their own cost: some concise but correct intermediate responses appeared semantically uncertain, triggering unnecessary reprompting that occasionally degraded an otherwise successful reasoning trajectory. Prior work has shown that confidence often corresponds with correctness~\citep{kuhn2023semanticuncertaintylinguisticinvariances}, but it does not guarantee it in either direction --- a response can be confidently wrong or hesitantly right.

A related limitation concerns the CEO's recovery policy once concern is correctly triggered. In several traces, the CEO responded to persistent semantic concern by repeatedly reprompting the same worker with only minor phrasing changes, rather than switching workers, forcing multi-agent delegation, or invoking a different reasoning strategy. In one MMLU trace, this pattern consumed the full worker-round budget on a tangential sub-question before the reprompt cap forced termination, without the CEO ever returning to the original question. Thus, semantic uncertainty's value depends not only on whether instability is detected, but by whether the CEO's recovery policy diversifies its response. Detecting a problem correctly is necessary but not sufficient if the corrective action available to the CEO is itself narrow.

A further limitation concerns evaluation scale. Due to compute and time constraints, benchmark subsets ranged from 100 to 500 questions rather than covering full benchmarks, and most configurations were run once rather than averaged over multiple seeds; the confidence intervals reported in Table~\ref{tab:results} partially address this but do not eliminate it. While these subsets were sufficient to reveal consistent directional trends across benchmarks, larger-scale evaluation --- particularly on the scope benchmarks, where effect sizes were smaller and more mixed --- would strengthen confidence in the magnitude of the reported gaps. Future work should extend evaluation to full benchmark splits and multiple random seeds per configuration.

Future work could combine semantic uncertainty with richer reasoning-quality estimates, such as trajectory-level uncertainty~\citep{duan2025upropinvestigatinguncertaintypropagation, zhao2024saupsituationawarenessuncertainty}, to help the CEO distinguish recoverable from unrecoverable failures. Adaptive thresholds --- replacing HASSUM's fixed values with ones tuned to task domain or worker capability --- are a related avenue, given that the model-size ablation already shows fixed thresholds behave differently across worker strength. A further direction is reintroducing HASHIRU's resource-budget system alongside semantic uncertainty rather than de-emphasizing it as was done in this paper: since reprompting and multi-agent delegation both carry a compute cost, a budget-aware policy could weigh the expected benefit of further refinement against its cost, particularly on benchmarks like the scope set where refinement is not reliably corrective.

More broadly, as agentic systems move toward persistent autonomy, recognizing low-confidence states and escalating accordingly becomes central to safe deployment. HASSUM demonstrates one practical step toward this by integrating semantic uncertainty directly into orchestration rather than as post-hoc analysis, and by showing that this signal's value depends on whether a task's dominant failure mode is one semantic instability can actually detect. This distinction --- uncertainty as necessary but not sufficient for recoverability --- extends beyond HASSUM's specific implementation: it suggests that future AI harnesses will likely require multiple layers of uncertainty-aware control, combining semantic confidence, factual verification, symbolic validation, and human escalation, since no single signal is well suited to every class of failure. Building such systems, and understanding which signal is appropriate for which failure mode, will be essential for making autonomous multi-agent systems reliable enough for real-world use.

\section{Conclusion}
As agentic systems move toward persistent autonomy, coping with low-confidence states becomes central to safe deployment. HASSUM demonstrates one practical step toward this goal by integrating semantic uncertainty directly into orchestration, and by showing that its value depends on whether a task's dominant failure mode is one semantic uncertainty can actually detect. This distinction --- uncertainty as necessary but not sufficient for recoverability --- extends beyond HASSUM: it suggests that future AI harnesses will likely require multiple layers of uncertainty-aware control, combining semantic confidence, factual verification, symbolic validation, and human escalation. Building such systems, and understanding which signal is appropriate for which failure mode, will be essential for making autonomous multi-agent systems reliable enough for real-world use.

\newpage
\bibliographystyle{plainnat}
\bibliography{references}

\newpage

\appendix
\section{Details of Metric Isolation Ablation}
\label{appendix:A}

The following execution traces and figures provide additional detail supporting the ablation results discussed in Section~\ref{sec:ablation}.

\begin{tcolorbox}[float, floatplacement=!h, title={\textbf{Summary 1$a$: Density only Run SQA Q3}}, colback=gray!5!white, colframe=gray!75!black]
\fontsize{10pt}{8pt}\selectfont
\label{box:1c}
In the density-only ablation, the CEO delegated the task to the BiographyResearcher agent. During the first round, the worker was asked a broad comparison question: whether the University of Pittsburgh founder and Judith Sheindlin had a great deal in common. The BiographyResearcher agent responded by incorrectly identifying Andrew Mellon as the founder and concluded that he and Judith Sheindlin did not have much in common. The response focused on differences in profession, historical period, and public role, leading to a clear ``no'' conclusion. This first response produced a semantic density of 0.6489, which was below the threshold of 0.8 and therefore triggered a semantic quality concern. Because density was below threshold, the CEO issued a follow-up prompt to the same BiographyResearcher agent. This second prompt simplified the task and explicitly asked for a yes-or-no answer with a brief explanation. However, the worker again returned ``No,'' this time still relying on incorrect assumptions about the founder and failing to recover the correct reasoning path. The second response produced a semantic density of 0.8032, which was just above the threshold, so the CEO accepted the answer and terminated the process. This run shows that density alone was able to detect that the first response was unstable enough to justify a reprompt, but the refinement strategy did not meaningfully improve reasoning quality. Once the second answer crossed the density threshold, the system stopped despite still being factually incorrect.
\end{tcolorbox}

\begin{tcolorbox}[float, floatplacement=!h, title={\textbf{Summary 1$b$: Entropy only Run SQA Q3}}, colback=gray!5!white, colframe=gray!75!black]
\fontsize{10pt}{8pt}\selectfont
\label{box:1d}
In the entropy-only ablation, the CEO again relied on the BiographyResearcher agent, but the interaction structure was different because semantic density was disabled. Instead of asking for a direct comparison first, the CEO decomposed the problem into two separate retrieval steps. In the first round, the worker was asked to provide information about the founder of the University of Pittsburgh. The agent again incorrectly identified Andrew Mellon as the founder and returned a concise biography describing him as a wealthy financier, philanthropist, and institutional founder. The semantic entropy for this response was 1.0094, which remained below the threshold of 1.1, so no concern was raised. In the second round, the worker was asked to provide information about Judith Sheindlin. The agent returned a detailed biography describing her as a lawyer, academic, and television personality. This response produced an even lower entropy value of 1.0094, again below threshold and therefore considered acceptable. The CEO then synthesized these two biographies and concluded that the two individuals did not have much in common, producing the final answer ``no.'' Since both responses were internally consistent and semantically stable, entropy never triggered concern. However, the core factual assumption was still wrong because the worker had misidentified the founder from the beginning. This run demonstrates an important limitation of entropy-only reasoning. Semantic entropy can detect instability across sampled meanings, but it cannot detect confidently wrong factual premises when the model remains internally consistent.
\end{tcolorbox}
\newpage
\section{Details of Threshold Ablation}
\label{appendix:B}

\begin{figure}[h!]
\centering
\includegraphics[width=0.7\linewidth]{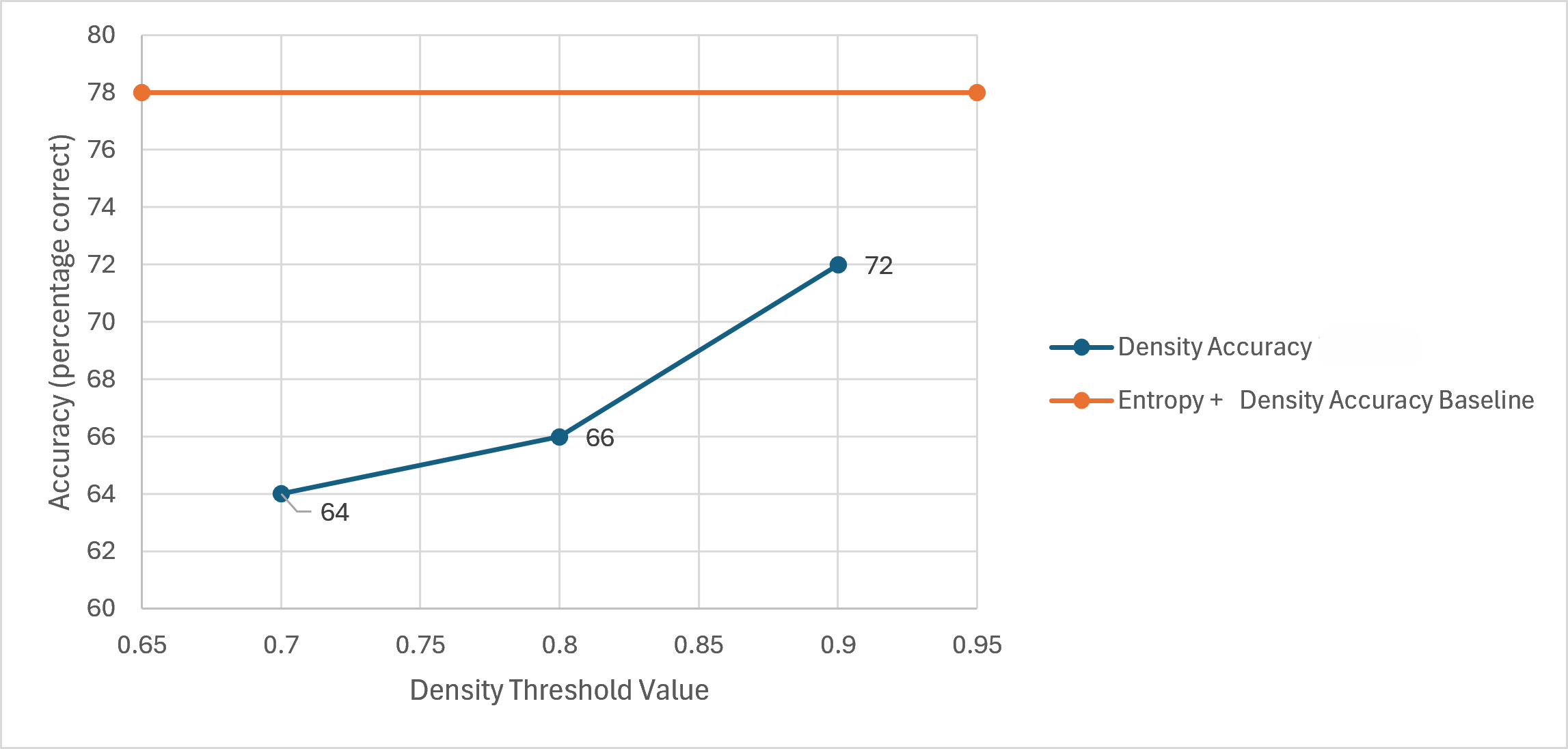}
\caption{Accuracy between different thresholds for acceptable semantic density. The y-values are the percentage of correctly answered questions at a specific threshold indicated by the x-values. The orange line is a baseline accuracy from the run with semantic density at 0.8 and semantic entropy at 1.1. The higher the threshold value, the better the performance since the CEO becomes stricter with a higher threshold and is more likely to reprompt agents.}
\label{fg:ablations_SD_threshold}
\end{figure}

\begin{figure}[h!]
\centering
\includegraphics[width=0.8\linewidth]{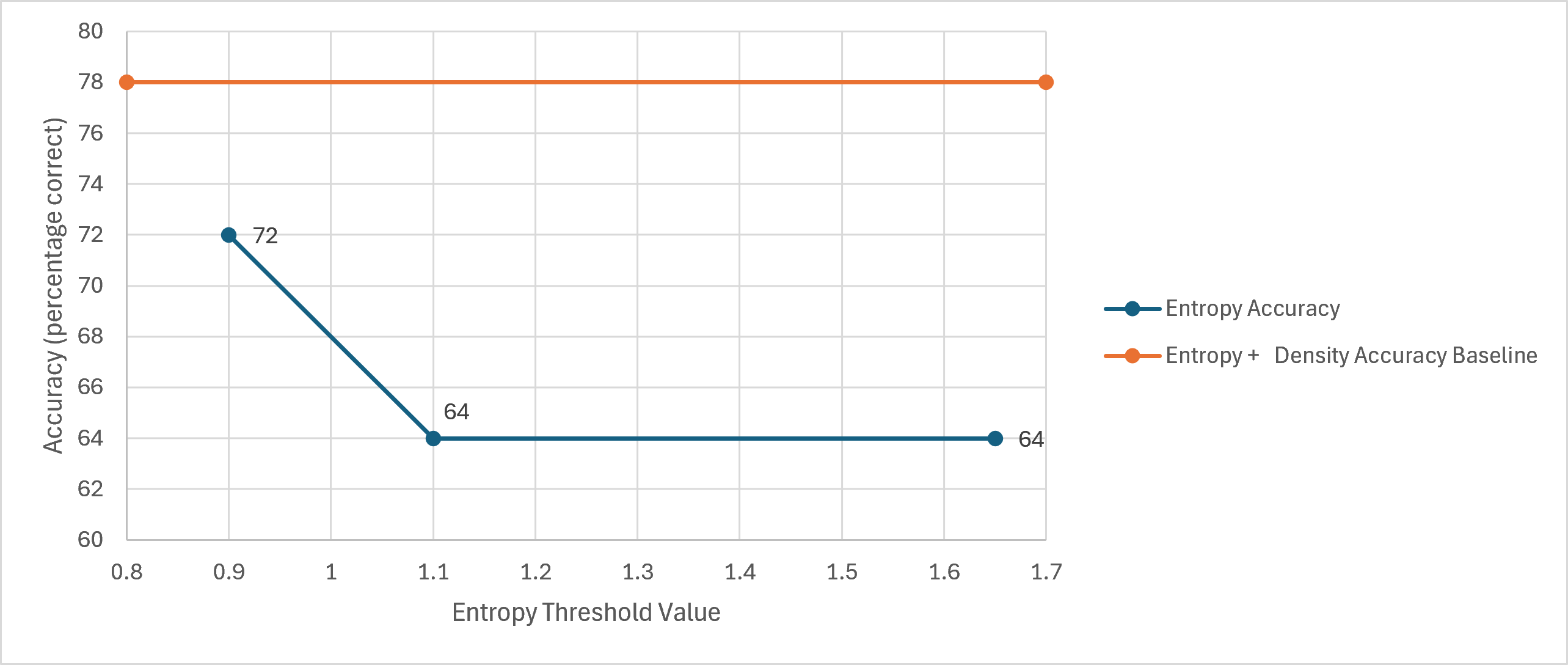}
\caption{Accuracy between different thresholds for acceptable semantic entropy. The y-values are the percentage of correctly answered questions at a specific threshold indicated by the x-values. The orange line is a baseline accuracy from using density at 0.8 and entropy at 1.1. The broad trend is that the lower the threshold value, the better the performance is, since the CEO becomes stricter the lower the threshold and is more likely to reprompt agents.}
\label{fg:ablations_SE_threshold}
\end{figure}

Figures~\ref{fg:ablations_SD_threshold} and~\ref{fg:ablations_SE_threshold} show the accuracy of HASSUM for different semantic density and semantic entropy thresholds discussed in Section~\ref{sec:ablation}. 

\newpage
\section{Details of Worker Capability Ablation}
\label{appendix:C}

Below are more details on the runs of HASSUM with smaller models versus GPT-5.4 in StrategyQA. Summaries 2a and 2b contain high level descriptions of the example runs. Both runs answer the same question:
\begin{displayquote}
    \emph{Was Kurt Cobain's death indirectly caused by Daniel LeFever?}
\end{displayquote}
Overall accuracy in the StrategyQA domain is shown in Table~\ref{tab:small_vs_SOTA_accuracy}.

\begin{tcolorbox}[float, floatplacement=!h, title={\textbf{Summary 2$a$: Smaller Local Worker Models}}, colback=gray!5!white, colframe=gray!75!black]
\fontsize{10pt}{8pt}\selectfont
\label{box:4a1}
In Run A, the CEO began by assigning the task to the BiographyResearcher agent. The initial prompt asked whether Daniel LeFever was directly or indirectly responsible for Kurt Cobain's death and requested a definitive yes-or-no answer. The worker answered ``No,'' but the response produced a semantic density of only 0.5574, well below the threshold of 0.8, and an entropy of 1.0549. This triggered semantic concern and caused the CEO to continue investigating. The CEO reprompted the same BiographyResearcher agent with a more constrained prompt emphasizing verifiable facts and avoidance of speculation. The worker again answered ``No,'' but density fell even further to 0.4250, keeping concern active. The CEO then switched to the HistoryExpert agent, which also answered ``No,'' but density remained low at 0.4304. Because repeated single-agent calls still produced low semantic confidence, the CEO escalated to multi-agent reasoning using AskMultipleAgents. Both BiographyResearcher and HistoryExpert were asked to independently analyze the relationship between Daniel LeFever and Kurt Cobain. Although both agents again converged on ``No,'' the responses contained factual inconsistencies and hallucinated details, such as incorrectly describing Daniel LeFever as Nirvana's drummer or associating him with unrelated bands. Density improved slightly but remained below threshold, with values of 0.7078 and 0.7238 across the multi-agent rounds. Only after six total worker-tool completions did the CEO finally receive a response with acceptable semantic quality. In the final round, the HistoryExpert returned another ``No'' answer with semantic density of 1.0 and entropy of 0.0. At this point the CEO accepted the answer and finalized the response. This run required six worker rounds, five semantic concern flags, and two explicit reprompts before the CEO trusted the result. The mean semantic density across finish rows was only 0.6407, while mean entropy remained relatively high at 0.6783. The system was forced to spend significant additional computation simply to gain confidence in an answer that was already being repeated from the beginning.
\vspace*{-2ex}
\end{tcolorbox}

\begin{tcolorbox}[float, floatplacement=!h, title={\textbf{Summary 2$b$: GPT-5.4 Worker Agents}}, colback=gray!5!white, colframe=gray!75!black]
\fontsize{10pt}{8pt}\selectfont
\label{box:4a2}
In Run B, the CEO used the GPT-5.4-based BiographyResearcherPlus agent. The initial prompt was much simpler: it directly asked whether Kurt Cobain's death was indirectly caused by Daniel LeFever and requested a definitive yes-or-no answer. The worker immediately responded: ``No. There is no credible evidence that Daniel LeFever indirectly caused Kurt Cobain's death.'' Unlike Run A, this response produced extremely strong semantic metrics on the first attempt. Semantic density was 0.9715, far above threshold, while semantic entropy was only 0.0523, indicating extremely high consistency and confidence across semantic samples. Because neither metric triggered concern, the CEO did not reprompt, did not invoke additional workers, and did not use multi-agent decomposition. The final answer was accepted after a single worker call. This run required only one worker-tool completion, zero reprompts, and zero semantic concern flags. The CEO was able to terminate immediately because the stronger worker model produced an answer that was both semantically concentrated and highly stable from the start.
\end{tcolorbox}

\begin{table}[h!]
\centering
\begin{tabular}{lc}
\hline
Model Type & Result (\%) \\
\hline
Llama-3.2 and Deepseek-R1 & 78 \\
GPT-5.4 & 78 \\
\hline
\end{tabular}
\caption{Two ablation runs that compare the performance of HASSUM using smaller models (Llama-3.2, Deepseek-R1) versus HASSUM using a larger model (GPT-5.4) as workers in the StrategyQA benchmark. The values in the table are the percentage of questions that each run answered correctly. Both runs had the same accuracy, showing how HASSUM can use semantic uncertainty to compensate for weaker models.}
\label{tab:small_vs_SOTA_accuracy}
\end{table}

\newpage
\section{Likelihood-Aware Semantic Entropy}
\label{appendix:D}

HASSUM optionally extends semantic entropy to incorporate sequence log-probabilities, following the likelihood-aware formulations proposed in follow-up semantic entropy work~\citep{Farquhar_Kossen_Kuhn_Gal_2024}. When supported local Ollama models are used, the orchestrator collects token-level log-probabilities for each sampled completion and passes the resulting sequence likelihoods to the semantic uncertainty backend. In addition to the standard cluster-assignment entropy computed from empirical semantic cluster frequencies, the backend can compute likelihood-aware variants, such as Rao--Blackwellized and predictive semantic entropy, which weight semantic clusters according to the generative model's estimated likelihood of each sampled response rather than treating all samples equally~\citep{Farquhar_Kossen_Kuhn_Gal_2024}. Semantic density remains based only on embedding-space similarity and is not extended with likelihood weighting. In the HASSUM implementation, the entropy value is returned along with auxiliary information such as cluster assignments and sample counts. The quality of entropy estimates depends heavily on the semantic clustering procedure; threshold selection and the choice of entailment model can significantly affect results. Similarly, the quality of density estimates depends on the embedding model used to represent responses, and the choice of similarity function influences the sensitivity of the density estimate. Both thresholds used in the main experiments were derived empirically, as no obvious heuristic was available to adjust them based on task or model type.

\end{document}